\documentclass[runningheads]{llncs}

\usepackage[final,year=2026,ID=8830]{eccv}

\usepackage[table]{xcolor}
\usepackage{arydshln}
\usepackage{booktabs}
\usepackage{amssymb}
\usepackage{amsmath}
\usepackage{multirow}
\usepackage{graphicx}
\usepackage{float}
\usepackage{bm}
\usepackage{pifont}
\usepackage{wrapfig}
\usepackage[ruled,linesnumbered,noline]{algorithm2e}
\usepackage[most]{tcolorbox}
\usepackage{enumitem}
\usepackage{courier}
\usepackage{eccvabbrv}

\newcommand{\cmark}{\ding{51}}
\newcommand{\xmark}{\ding{55}}

\definecolor{promptblue}{RGB}{40,88,160}
\definecolor{promptgray}{RGB}{245,246,248}
\definecolor{promptborder}{RGB}{210,214,220}
\definecolor{first}{HTML}{547CB1}
\definecolor{improve}{HTML}{1E73C4}

\newcommand{\upcolor}[1]{\textcolor{first}{#1}}

\newtcblisting{promptbox}{
  enhanced,
  breakable,
  colback=promptgray,
  colframe=promptborder,
  boxrule=0.6pt,
  arc=2mm,
  left=2mm,
  right=2mm,
  top=1mm,
  bottom=1mm,
  listing only,
  listing options={
    basicstyle=\ttfamily\small,
    breaklines=true,
    columns=fullflexible
  }
}

\usepackage{hyperref}
\usepackage{orcidlink}

\usepackage{hyperref}

\usepackage{orcidlink}

\begin{document}

\title{A$^2$BFR: Attribute-Aware Blind Face Restoration} 


\author{Chenxin Zhu\inst{1} \and
Yushun Fang\inst{1,2} \and
Lu Liu\inst{1} \and
Shibo Yin\inst{2} \and
Xiaohong Liu\inst{1} \and
Xiaoyun Zhang\inst{1} \and
Qiang Hu\inst{1} \and
Guangtao Zhai\inst{1} 
}

\authorrunning{C. Zhu et al.}

\institute{
Shanghai Jiao Tong University, Shanghai, China
\and
Xiaohongshu Inc., Shanghai, China
}

\maketitle



\begin{figure}[H]
    \centering
    \vspace{-20pt}
    \includegraphics[width=1.0\linewidth]{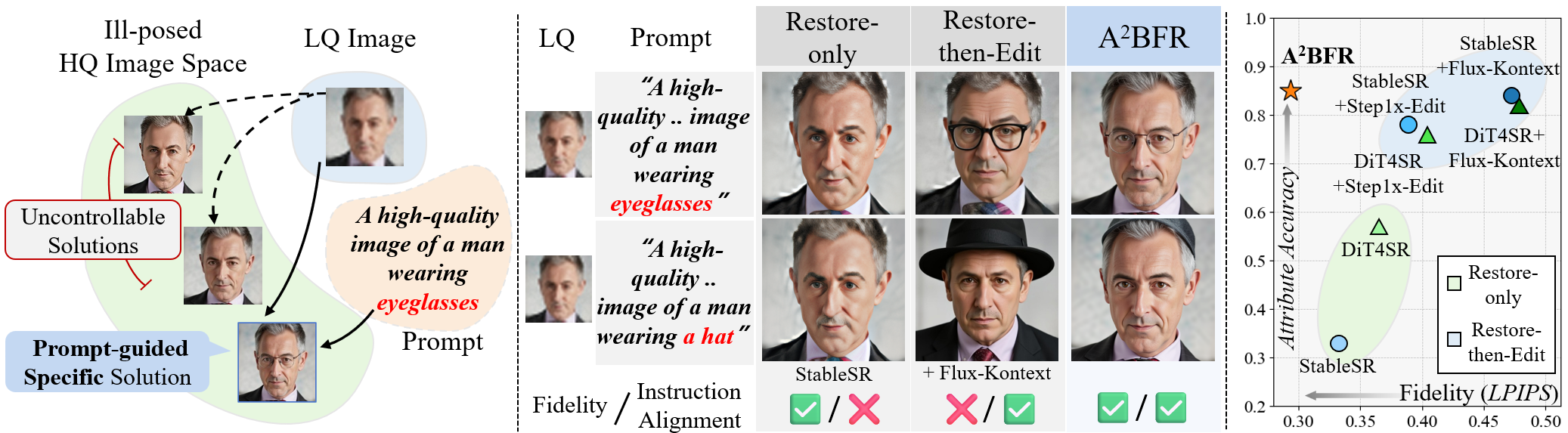}
    \caption{\textbf{Left:} Prompt-guided specific restoration avoids uncontrolled solutions in ill-posed BFR.
\textbf{Middle:} Compared with restore-only and restore-then-edit pipelines, A$^\text{2}$BFR achieves faithful yet attribute-aligned restoration.
\textbf{Right:} Quantitative results show that A$^\text{2}$BFR simultaneously achieves superior fidelity and attribute accuracy, outperforming all compared restoration and editing methods.}
    \label{fig:teaser}
    \vspace{-16pt}
\end{figure}

\begin{abstract}

Blind face restoration (BFR) aims to recover high-quality facial images from degraded inputs, yet its inherently ill-posed nature leads to ambiguous and uncontrollable solutions. Recent diffusion-based BFR methods improve perceptual quality but remain uncontrollable, whereas text-guided face editing enables attribute manipulation without reliable restoration.  To address these issues,  we propose A$^2$BFR, an attribute-aware blind face restoration framework that unifies high-fidelity reconstruction with prompt-controllable generation. Built upon a Diffusion Transformer backbone with unified image–text cross-modal attention, A$^2$BFR jointly conditions the denoising trajectory on both degraded inputs and textual prompts. To inject semantic priors, we introduce attribute-aware learning, which supervises denoising latents using facial attribute embeddings extracted by an attribute-aware encoder. To further enhance prompt controllability, we introduce semantic dual-training, which leverages the pairwise attribute variations in our newly curated AttrFace-90K dataset to enforce attribute discrimination while preserving fidelity. Extensive experiments demonstrate that A$^2$BFR achieves state-of-the-art performance in both restoration fidelity and instruction adherence, outperforming diffusion-based BFR baselines by \textbf{-0.0467} LPIPS and \textbf{+52.58\%} attribute accuracy, while enabling fine-grained, prompt-controllable restoration even under severe degradations.


\end{abstract}    
\section{Introduction}
\label{sec:intro}

Blind face restoration (BFR)~\cite{GFPGAN13} aims to recover a high-quality facial image from a degraded input affected by blur~\cite{GFPGAN71}, noise~\cite{GFPGAN39}, downsampling~\cite{GFPGAN58}, or compression~\cite{GFPGAN12}. The problem is inherently ill-posed~\cite{lugmayr2021ntire,gandikota2024text}: many plausible high-quality faces may correspond to the same degraded observation, especially when semantic cues such as eyeglasses, hairstyle, or expression are corrupted or missing. Consequently, without explicit constraints, BFR models may converge to arbitrary solutions in the high-quality image space. In contrast, prompt-guided restoration mitigates this ambiguity by introducing semantic constraints, thereby avoiding uncontrolled solutions under ill-posed BFR (Fig.~\ref{fig:teaser}, Left).

Recent diffusion-based BFR methods exploit strong generative priors to enhance perceptual realism~\cite{yue2022difface,lin2023diffbir,wang2023dr2,wang2024exploiting}, yet they remain largely uncontrollable. Most methods treat text prompts as optional or fixed inputs, relying almost entirely on degraded visual cues, which leads to inconsistent or unintended attribute restoration when key semantics are missing. Several recent works~\cite{duan2025dit4sr,liu2024improved,tao2024overcoming,yu2024scaling} attempt to incorporate semantic priors via recognition models or Vision-Language Models (VLMs), but text is still treated as a weak auxiliary signal rather than a primary control mechanism. Consequently, such 
restore-only designs often fail to follow user-specified attribute instructions. A straightforward alternative is the restore-then-edit pipeline, which can enforce prompts but frequently compromises fidelity, inducing identity shifts or structural drift (Fig.~\ref{fig:teaser}, Middle).

The limited prompt adherence of restore-only methods is largely driven by a supervision imbalance. 
In many BFR pipelines, prompts are either fixed or automatically generated by VLMs, which are often low-quality and weakly correlated with the ground-truth (GT) target. 
Meanwhile, low-quality (LQ) input remains strongly correlated with GT,  
and consequently, optimization is dominated by the LQ-GT mapping and the model learns to treat text as a secondary cue, leading to poor prompt adherence. 
This suggests two requirements for controllable BFR: (i) an attribute-centered dataset with explicit paired attribute supervision, and (ii) a prompt-centered training strategy that elevates text to an explicit constraint during restoration.

In this work, we propose \textbf{A$^\text{2}$BFR}, an attribute-aware blind face restoration framework that unifies high-fidelity reconstruction with prompt-controllable generation. Built upon a rectified-flow Diffusion Transformer (DiT) with unified image-text attention, A$^\text{2}$BFR treats textual instructions as a primary control signal that complements degraded visual cues. This joint conditioning establishes a prompt-guided restoration space, enabling the denoising trajectory to follow target semantics when the low-quality input lacks discriminative attributes.

Beyond architectural conditioning, A$^\text{2}$BFR introduces two key mechanisms.
First, attribute-aware learning (AAL) supervises denoising latents using attribute embeddings extracted by an attribute-aware encoder. By enforcing semantic consistency between intermediate predictions and ground-truth attributes, AAL prevents semantic drift, and preserves fine-grained attribute details.
Second, semantic dual-training (SDT) leverages pairwise attribute variations in our curated dataset by associating each degraded input with two high-quality targets representing distinct attributes (e.g., with glasses vs. without glasses). This dual supervision encourages a discriminative attribute embedding space that supports fine-grained and prompt-controlled restoration while maintaining shared identity and geometry. Together, AAL and SDT enable A$^\text{2}$BFR to outperform restore-only and restore-then-edit pipelines in both fidelity and attribute alignment (Fig.~\ref{fig:teaser}, Right).

To support controllable BFR, we construct \textbf{AttrFace-90K}, a large-scale dataset containing 
90K HQ image pairs and 180K text descriptions across 12 attribute categories. Compared with prior face-editing datasets~\cite{zhu2025seed,shao2022detecting}, AttrFace-90K provides paired facial images with rich attribute diversity, strong identity preservation, and high structural consistency, establishing the first dataset explicitly tailored to attribute-aware blind face restoration. Extensive experiments show that A$^\text{2}$BFR achieves state-of-the-art performance in both perceptual quality and attribute adherence, improving LPIPS by -0.0467 and Attribute Accuracy by +52.58\% over diffusion-based BFR baselines.

Our main contributions are summarized as follows: 
\begin{itemize}
    \item We propose \textbf{A$^\text{2}$BFR}, the first framework that unifies high-fidelity blind face restoration with prompt-controllable attribute editing, enabling user-customizable reconstructions without sacrificing fidelity.
    \item We introduce \textbf{Semantic Dual-Training}, which leverages face-editing data to enforce prompt-dependent separation for Blind Face Restoration; together with \textbf{Attribute-Aware Learning}, it improves attribute alignment and controllability while preserving identity and facial geometry.
    \item We construct \textbf{AttrFace-90K}, a large-scale dataset containing over 90K facial pairs and 180K text annotations across 12 attributes, designed for controllable and semantically aligned BFR.
    \item Extensive experiments demonstrate that A$^\text{2}$BFR achieves SOTA performance across multiple datasets, reducing LPIPS by \textbf{0.0467} and improving attribute accuracy by \textbf{52.58\%} compared with diffusion-based BFR baselines.
\end{itemize}

\section{Related Work}
\label{sec: Related Work}

\noindent\textbf{Blind Face Restoration.}
Blind Face Restoration aims to recover HQ facial images from LQ inputs degraded by unknown and complex processes.
Recent methods alleviate the ill-posed nature of this task by incorporating external priors, including geometric priors~\cite{chen2018fsrnet,kim2019progressive}, vector-quantized (VQ) codebook priors~\cite{zhou2022towards,gu2022vqfr}, and generative priors.
Generative priors, owing to their strong texture synthesis ability, have evolved from GAN-based~\cite{yang2021gan,wang2021gfpgan} to diffusion-based approaches~\cite{yue2022difface,lin2023diffbir,wang2023dr2,wang2024exploiting, zhang2025td, fang2025robust, wang2025osdface}.
With the rise of multimodal learning, frameworks such as StableSR~\cite{wang2024exploiting} and DiT4SR~\cite{duan2025dit4sr} integrate text-to-image diffusion priors to enhance perceptual realism.
StableSR employs fixed template prompts based on Stable Diffusion v2.1, whereas DiT4SR leverages LLaVA~\cite{liu2024improved} to automatically generate descriptive annotations.
However, these prompt-based frameworks mainly improve perceptual quality but lack controllability, often producing stochastic or inconsistent outputs that cannot be regulated by semantic instructions.
This motivates a controllable BFR framework that achieves semantically aligned, text-conditioned restoration.

\smallskip
\noindent\textbf{Image Editing.}
Image editing methods can be broadly categorized into training-free and training-based approaches.
Training-free techniques manipulate the diffusion denoising process via latent inversion or attention control~\cite{tumanyan2023plug,rombach2022high,mokady2023null,cao2023masactrl,wang2024taming}, whereas training-based methods fine-tune diffusion models with synthetic or VLM-augmented supervision~\cite{brooks2023instructpix2pix,fu2023guiding,sheynin2024emu,liu2025fbench}.
For face editing, StyleCLIP and ChatFace~\cite{patashnik2021styleclip,yue2023chatface} enable text-driven control but often suffer from attribute entanglement and identity drift.
To enhance structural alignment, recent works introduce geometric conditions such as masks, landmarks, or layouts~\cite{zhang2025museface,mofayezi2024m,li2022towards,sun2024lafs}.
Similarly, DiT-based models rasterize landmarks to guide diffusion~\cite{tan2024ominicontrol,pan2025transfer}, yet pixel-level conditioning may cause template-copying artifacts and high computational cost.
When directly applied to BFR outputs, such editing pipelines fail to maintain semantic consistency with degraded inputs, underscoring the need for a unified, controllable, and attribute-aware BFR framework.

\section{Dataset: AttrFace-90K}

\textbf{Motivation.}
High-quality paired data are indispensable for training controllable face restoration models. 
However, existing facial editing datasets such as SeqDeepFake~\cite{shao2022detecting} and SEED~\cite{zhu2025seed} suffer from two key limitations: 
(1) limited attribute diversity, offering only coarse categorical labels; and 
(2) degraded image fidelity and quality caused by outdated synthesis pipelines. 
To address these issues, we construct AttrFace-90K, a large-scale and high-fidelity dataset tailored for attribute-aware blind face restoration. 
It provides 90K source-target facial image pairs and 180K fine-grained captions covering 12 attributes (e.g., age, hairstyle, and expression), 
substantially surpassing existing datasets in both scale and semantic richness, as shown in Table~\ref{tab:attrface_comparison}. 
\begin{table}[tbp]
  \centering
  \caption{Comparison of facial editing datasets.
AttrFace-90K provides the largest scale, the most diverse attributes, and the most powerful construction pipeline among existing datasets.}
  \resizebox{\columnwidth}{!}{
  \vspace{-15pt}  
    \begin{tabular}{lccccc}  
      \toprule[1pt]
      \textbf{ Datasets} & \textbf{ Size} & \textbf{ Attributes} & \textbf{Fine-grained Captions} & \textbf{ Construction Pipeline}  \\
      \midrule
        SeqDeepFake~\cite{shao2022detecting}  & 49,920  & 5 & \xmark & Talk-to-Edit \\
        SEED~\cite{zhu2025seed}  & 91,526  & 6 & \xmark & UltraEdit, SDXL... \\
      \hdashline
      \rowcolor{gray!20} 
       \textbf{AttrFace-90K} & \textbf{180,424} & \textbf{12} & \cmark & \textbf{FlowEdit + SD3.5} \\
      \bottomrule[1pt]
    \end{tabular}
  
  \label{tab:attrface_comparison}
  }
\end{table}

  

\subsection{Dataset Construction}
\begin{figure*}[tbp]
    \centering
    \includegraphics[width=\linewidth]{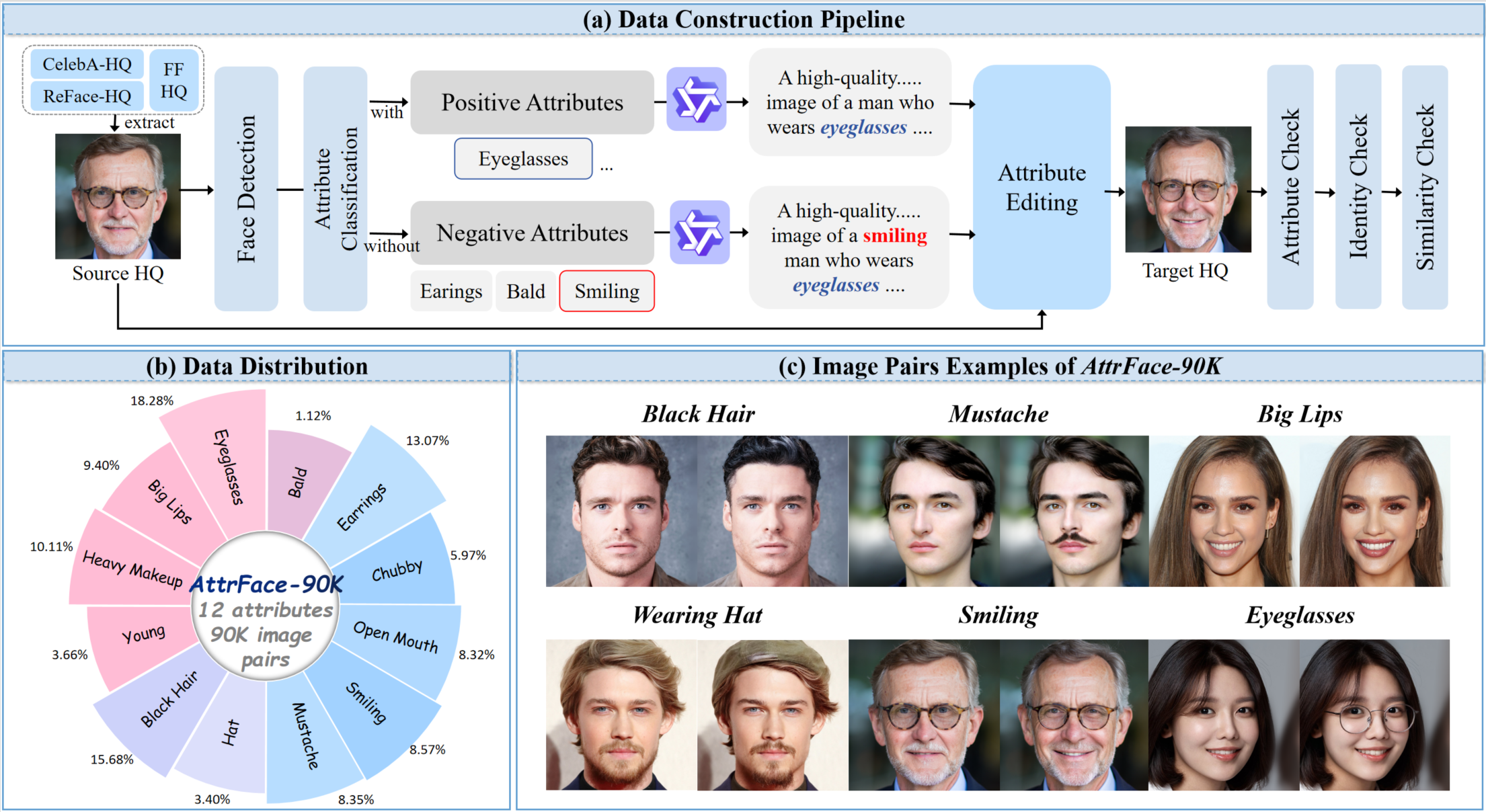}
    \caption{AttrFace-90K overview.
  (a) Data construction pipeline:
  starting from HQ faces (FFHQ/ReFace-HQ/CelebA-HQ), we extract features, classify attributes, build prompts with Qwen3, and perform attribute editing with FlowEdit+SD3.5, followed by post-checks.
  (b) Attribute distribution: percentage of each controllable attribute across 90K pairs.
  (c) Image pair examples: source–target pairs for typical attributes.}
\vspace{-10pt}
    \label{fig:lora}
\end{figure*}

\noindent \textbf{Prompt Construction.}
We collect over 120K high-quality face images from FFHQ~\cite{karras2019style}, ReFace-HQ~\cite{tao2024overcoming}, and CelebA-HQ~\cite{karras2017progressive}.  
Each image is detected and cropped using RetinaFace~\cite{deng2020retinaface}, and its 12 facial attributes are predicted by the FaRL recognizer~\cite{zheng2022general}. 
Following~\cite{tao2024overcoming}, attributes with confidence above $0.6$ are regarded as positive and those below $0.4$ as negative. 
As demonstrated in Fig.~\ref{fig:lora}(a), we employ the large language model Qwen3~\cite{yang2025qwen3technicalreport} to verbalize these attributes into natural, human-like captions. 
Each image is paired with a source prompt which describes existing attributes and  target prompts modifying one attribute to a contrasting semantic state. 
This dual-prompt design directly supports the SDT scheme in our A$^\text{2}$BFR framework.

 \noindent \textbf{Attribute Editing.}
Using the constructed prompt pairs, we synthesize edited targets with FlowEdit~\cite{kulikov2025flowedit} combined with the diffusion model SD3.5. 
To mitigate over-editing and identity drift, we introduce a controllable noise-blending mechanism (provided in the \textit{supplementary materials}), which interpolates the original and edited noise maps to balance semantic modification and structural preservation. 
This adjustment enhances identity consistency while maintaining high-fidelity attribute editing, as shown in Fig.~\ref{fig:lora}(c). 

 \noindent \textbf{Quality Control.}
To ensure the reliability of our dataset, we employ a progressive quality-control pipeline: (i) an attribute check to verify that the target attribute edits are correctly applied, (ii) an identity check to filter out shifted identities during editing, and (iii) a similarity check to enforce structural preservation for better alignment with the BFR task. Beyond automated filtering, we conduct a rigorous manual review to ensure the highest data quality. These quality control strategies yield visually coherent, attribute-faithful, and identity-consistent results that are essential for controllable restoration training. Details of quality control will be provided in the \textit{supplementary materials}. 

\subsection{Dataset Statistics}
AttrFace-90K consists of 90,212 edited image pairs and 180,424 descriptive captions. 
After quality control, the final average identity similarity of AttrFace-90K reaches 0.6387, higher than 0.549 reported by existing identity-preserving face restoration work~\cite{fang2025robust}, indicating strong identity preservation. The attribute frequency distribution is visualized in Fig.~\ref{fig:lora}(b), showing balanced coverage across age, expression, and appearance variations. AttrFace-90K thus establishes the first large-scale dataset enabling fine-grained, prompt-controllable, and identity-preserving face restoration.

\section{Method}
\begin{figure*}[tbp]
    \centering
    \includegraphics[width=\linewidth]{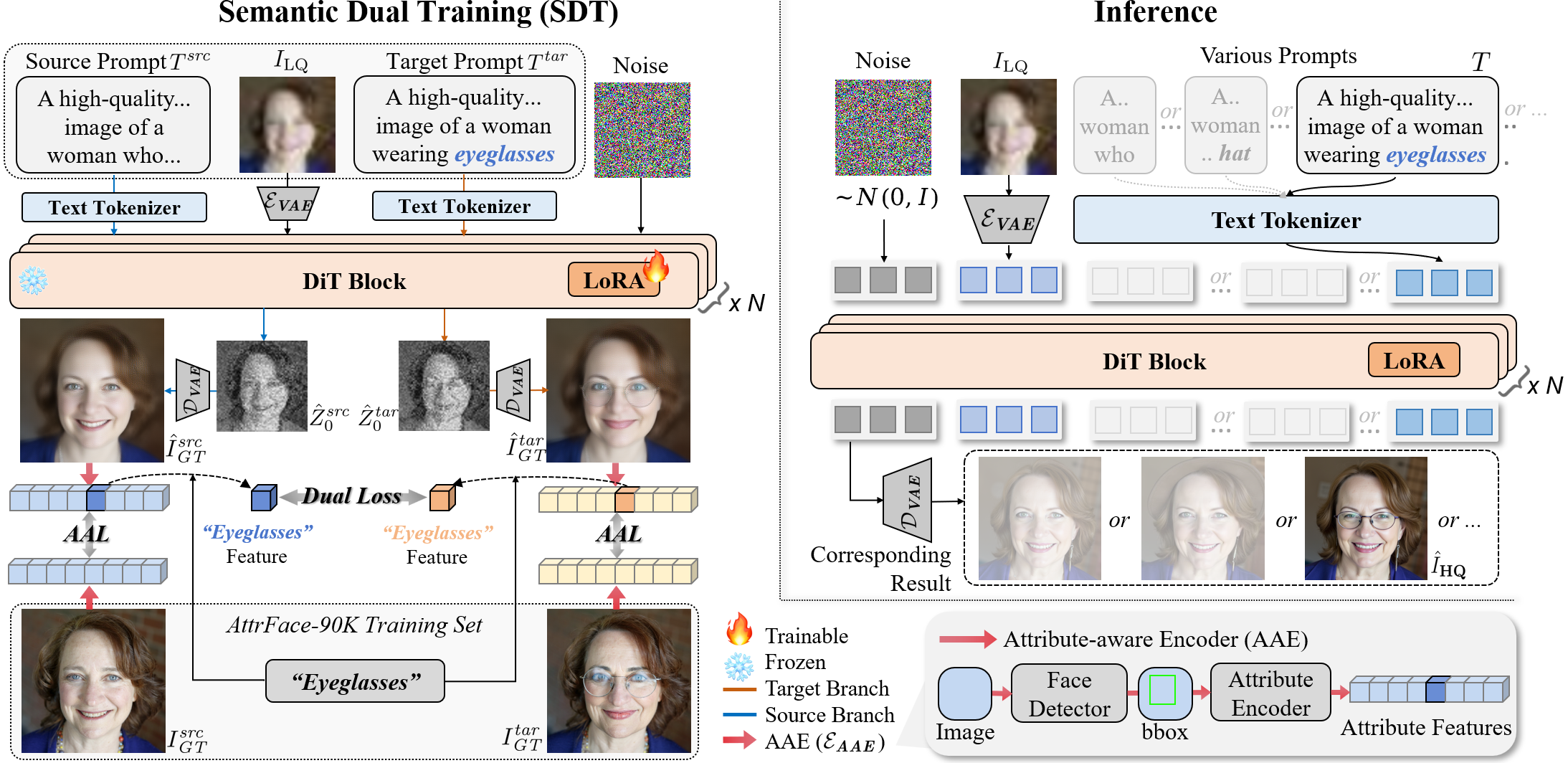}
    \caption{Overview of our $\text{A}^\text{2}$BFR. 
    \textbf{Left:} The semantic dual-training (SDT) strategy pairs a single LQ input with two GT images driven by different attribute prompts, and enforces maximal separation between their latent representations in the semantic space. This encourages the model to learn attribute-discriminative, prompt-aligned behaviors.  \textbf{Right:} At inference time, $\text{A}^\text{2}$BFR takes an LQ image and a user-specified prompt, and generates an attribute-aligned, high-quality restoration through denoising process. }
    \label{fig:framework}
\end{figure*}

\subsection{Overview}
\label{sec:overview}

As illustrated in Fig.~\ref{fig:framework}, A$^\text{2}$BFR enables attribute-aware, prompt-controllable blind face restoration. Given a degraded image $I_{\mathrm{LQ}}$ and a textual prompt $T$, the model produces a high-quality reconstruction $\hat I_{\mathrm{HQ}}$ that is visually faithful to the input and semantically aligned with the prompt.

A$^\text{2}$BFR is built upon a pretrained Diffusion Transformer backbone (Flux~\cite{flux2024}) with lightweight LoRA adapters for efficient fine-tuning. Inspired by ~\cite{tan2024ominicontrol}, we introduce a dual-branch conditioning scheme that injects LQ guidance while keeping the backbone weights frozen. Concretely, visual tokens from the noise and $I_{\mathrm{LQ}}$, together with textual tokens from $T$, are jointly processed through unified cross-modal attention layers over a concatenated token sequence. This design steers the denoising trajectory using both pixel-level evidence and explicit semantic instructions, while retaining the pretrained prior.

For facial attribute priors, a two-stage attribute-aware encoder is utilized ($\mathcal{E}_{\text{AAE}}$): a face detector~\cite{deng2020retinaface} first localizes and aligns facial regions, and an attribute encoder~\cite{zheng2022general} extracts the corresponding attribute embeddings. These embeddings provide semantic supervision for the AAL and SDT schemes described below.

\subsection{Attribute-Aware Learning}
\label{sec:aal}

Recovering fine-grained semantic details (e.g., eyeglasses, smile, hair color) from degraded inputs requires high-level guidance beyond pixel-wise losses.
We propose an attribute-aware learning strategy that regularizes the diffusion process via pretrained facial attribute priors, explicitly aligning restored faces with their semantic attributes.

Given a training pair $(I_{\mathrm{LQ}}, I_{\mathrm{GT}})$, we encode them into the VAE latent space, obtaining latents $Z_0$ for $I_{\mathrm{GT}}$ and latents $Z_{\mathrm{LQ}}$ for $I_{\mathrm{LQ}}$. 
We then sample Gaussian noise $\epsilon \sim \mathcal{N}(0,I)$ and construct $Z_t$ at timestep $t\in[0,1]$ by linearly interpolating between $Z_0$ and $\epsilon$.
Conditioned on $Z_{\mathrm{LQ}}$, the rectified-flow predictor $V_\theta$ estimates the velocity field and yields the denoised latent
\begin{equation}
\hat Z_0 = Z_t - t \cdot V_{\theta}\!\left(Z_t,\, t,\, Z_{\mathrm{LQ}}\right).
\label{eq:generate_z_fake}
\end{equation}
We decode $\hat Z_0$ back to the image space and extract attribute embeddings using an attribute-aware encoder. The attribute loss is defined as
\begin{equation}
\mathcal{L}_{\mathrm{attr}}
=
\mathrm{BCE}\!\left(
\mathcal{E}_{\mathrm{AAE}}\!\left(\mathcal{D}_{\mathrm{VAE}}(\hat Z_0)\right),
\mathcal{E}_{\mathrm{AAE}}\!\left(I_{\mathrm{GT}}\right)
\right),
\label{eq:L_attr}
\end{equation}
where $\mathcal{D}_{\mathrm{VAE}}(\cdot)$ denotes the VAE decoder and $\mathrm{BCE}(\cdot,\cdot)$ is the binary cross-entropy loss.
This semantic supervision anchors the denoising trajectory to an attribute-aware embedding space, enforcing that the restored face matches the target attributes implied by the prompt and ground truth.

Let $\mathcal{L}_{\mathrm{diff}}$ denote the standard diffusion reconstruction loss of the DiT backbone. 
The overall objective of AAL is defined as
\begin{equation}
\mathcal{L}_{\mathrm{AAL}} \;=\; \mathcal{L}_{\mathrm{diff}} \;+\; \lambda\,\mathcal{L}_{\mathrm{attr}},
\end{equation}
where $\lambda$ balances pixel-level fidelity and semantic alignment. 
In practice, AAL augments blind face restoration with explicit semantic guidance, steering the optimization beyond purely pixel-driven reconstruction toward recovering prompt-relevant facial attributes.

\subsection{Semantic Dual-Training}
\label{sec:sdt}
While AAL improves semantic alignment, it does not by itself guarantee controllability, i.e., producing distinct restorations under different textual prompts given the same degraded input. 
To this end, we exploit the pairwise structure of AttrFace-90K and propose semantic dual-training, which enforces prompt-dependent separation in the attribute space while preserving restoration fidelity.

For each degraded image $I_{\mathrm{LQ}}$, AttrFace-90K provides an attribute-varying HQ pair. 
We thus form two branches with targets and prompts $(I^{\mathrm{src}}_{\mathrm{GT}}, T^{\mathrm{src}})$ and $(I^{\mathrm{tar}}_{\mathrm{GT}}, T^{\mathrm{tar}})$ that differ in a single attribute $a$. 
Both branches share network parameters and are processed by A$^\text{2}$BFR, yielding denoised latents $(\hat Z^{\mathrm{src}}_0,\hat Z^{\mathrm{tar}}_0)$ and decoded predictions $(\hat I^{\mathrm{src}}_{\mathrm{GT}}, \hat I^{\mathrm{tar}}_{\mathrm{GT}})$. 
We then extract the confidence score of the edited attribute $a$ using the attribute-aware encoder:
\begin{equation}
\mathcal{A}_{\mathrm{src}} = \big[\mathcal{E}_{\mathrm{AAE}}\!\left(\hat I^{\mathrm{src}}_{\mathrm{GT}}\right)\big]_a, \qquad
\mathcal{A}_{\mathrm{tar}} = \big[\mathcal{E}_{\mathrm{AAE}}\!\left(\hat I^{\mathrm{tar}}_{\mathrm{GT}}\right)\big]_a,
\end{equation}
where $\mathcal{E}_{\mathrm{AAE}}(\cdot)$ outputs attribute confidence scores and $[\cdot]_a\in[0,1]$ denotes the score of the edited attribute $a$.

To discourage attribute ambiguity, we explicitly separate the two branches along the annotated attribute $a$. 
Specifically, we impose a margin-based contrastive objective:
\begin{equation}
\mathcal{L}_{\mathrm{dual}}
=
\max\!\Big(0,\; m - \big|\mathcal{A}_{\mathrm{tar}}-\mathcal{A}_{\mathrm{src}}\big|\Big),
\label{eq:dual_loss}
\end{equation}
where $m>0$ is a margin applied to the confidence gap of the edited attribute $a$. 
This constraint encourages prompt-conditioned diversity, while the shared degraded input anchors identity and facial geometry, thereby maintaining structural consistency across branches.

The final training objective integrates AAL on both branches and the dual loss:
\begin{equation}
\mathcal{L}_{\mathrm{total}}
=
\big(\mathcal{L}_{\mathrm{AAL}}\big)_{\mathrm{src}}
+
\big(\mathcal{L}_{\mathrm{AAL}}\big)_{\mathrm{tar}}
+
\alpha\,\mathcal{L}_{\mathrm{dual}},
\end{equation}
where $\alpha$ controls the strength of prompt-induced attribute separation. 
Intuitively, AAL teaches the model to recognize and align facial attributes, whereas SDT compels the model to distinguish and control them under different prompts. 
Together, they enable A$^\text{2}$BFR to achieve high-fidelity, attribute-consistent, and prompt-controllable face restoration.

\vspace{-3pt}
\section{Experiments}
\vspace{-2pt}

\subsection{Experimental setup}




\textbf{Datasets.}
 We curate 35,000 paired samples from the FFHQ-derived portion of AttrFace-90K for training, and 1,200 paired samples from the ReFace-HQ-derived portion for evaluation, denoted as \textit{AttrFace-90K-Test}. Additionally, we include 1,102 testing images from CelebRef-HQ-Test~\cite{li2022learning}, and three widely used real-world datasets: LFW-Test~\cite{huang2008labeled}, Wider-Test~\cite{yang2016wider}, and WebPhoto-Test~\cite{wang2021gfpgan}. 
All images are center-cropped, aligned, and normalized following standard facial preprocessing protocols~\cite{tsai2024dual, wang2025osdface}. During training and fine-tuning, LQ images are synthesized online using the standard degradation pipeline detailed in the \textit{supplementary material}.



\noindent\textbf{Implementation Details.}
All experiments are conducted on the NVIDIA RTX~A6000 GPU. 
We adopt Flux~1.0-Dev as the pretrained diffusion prior and fine-tune it via LoRA for 60K iterations using the Prodigy optimizer. 
The learning rate is set to~1.0 with a batch size of~1, LoRA rank of~64, and LoRA alpha of~4. 
Dropout probabilities are set to~0.01. 
Unless otherwise specified, inference is performed at $512\times512$ resolution with classifier-free guidance enabled.

\noindent\textbf{Metrics.}
We assess three aspects: (1)~restoration quality, (2)~restoration fidelity, and (3)~attribute alignment.
For perceptual quality, we use HyperIQA~\cite{hyperiqa}, TopIQ~\cite{topiq}, MANIQA~\cite{maniqa}, and MUSIQ~\cite{musiq}, which measure the overall no-reference visual quality of the restored images.
For fidelity, we report LPIPS~\cite{lpips}, FID~\cite{heusel2017gans}, and IDS, where IDS is the cosine similarity between ArcFace~\cite{deng2019arcface} identity embeddings extracted from the restored images and their ground-truth counterparts.
Following recent image-editing studies~\cite{liu2025step1x,zhang2025lato}, we assess attribute alignment using Semantic Consistency (SC) and Attribute Accuracy (AA), both automatically estimated by Qwen3-VL~\cite{yang2025qwen3technicalreport} via visual reasoning. SC measures text--image semantic agreement, whereas AA evaluates whether the target attributes are correctly manifested.

\subsection{Comparison with Restoration Methods}

\begin{table*}[t]
\caption{Quantitative results of restoration quality comparison on AttrFace-90K-Test, CelebRef-HQ-Test~\cite{li2022learning}, LFW-Test~\cite{huang2008labeled}, Wider-Test~\cite{yang2016wider}, and WebPhoto-Test~\cite{wang2021gfpgan}. \textbf{Best} and \underline{second best} performance are highlighted. \upcolor{$\uparrow$} means higher is better, \upcolor{$\downarrow$} means lower is better. $\spadesuit$ denotes reference-based fidelity metrics, while $\heartsuit$ represents no-reference quality metrics.} 
\centering
{
\resizebox{\linewidth}{!}{
\begin{tabular}{l c c c c c c c c c c}
\toprule
 Metrics/Methods & GFPGAN & CodeFormer & VQFR & \quad DR2 \quad\quad  & DifFace & DAEFR & VSPBFR & StableSR & DiT4SR & \textbf{$\text{A}^\text{2}$BFR} \\
 
 \midrule
  \rowcolor{gray!20}\multicolumn{11}{c}{\ { \textit{AttrFace-90K-Test}}}\\
\midrule

$\spadesuit$ IDS~\cite{deng2019arcface} \upcolor{$\uparrow$} & 0.3242 & \underline{0.3337} & 0.3030 & 0.2782 & 0.2852 & 0.3012 & 0.2995 & 0.3289 & 0.3048 & \textbf{0.3495} \\
$\spadesuit$ FID~\cite{heusel2017gans} \upcolor{$\downarrow$} & 20.62 & 20.31 & \underline{18.46} & 43.92 & 27.22 & 23.61 & 26.82 & 25.91 & \textbf{18.10} & 18.69 \\
 $\spadesuit$ LPIPS~\cite{lpips} \upcolor{$\downarrow$} & \underline{\ {0.3070}} & 0.3154 & 0.3123 & 0.3676 & 0.3398 & 0.2934 & 0.3585 & 0.3396 & 0.3442 & \textbf{\textbf{0.2679}} \\
 $\heartsuit$ HyperIQA~\cite{hyperiqa} \upcolor{$\uparrow$} & 0.7380 & 0.7160 & 0.7156 & 0.6419 & 0.6127 & 0.7365 & 0.6815 & 0.6953 & \underline{\ {0.7609}} & \textbf{\textbf{0.8084}} \\
 $\heartsuit$ TopIQ~\cite{topiq} \upcolor{$\uparrow$} & 0.7956 & 0.7740 & 0.7482 & 0.5929 & 0.5938 & 0.7760 & 0.7061 & 0.7393 & \underline{\ {0.8499}} & \textbf{\textbf{0.8748}} \\
 $\heartsuit$ MANIQA~\cite{maniqa} \upcolor{$\uparrow$} & 0.5679 & 0.5210 & 0.5256 & 0.4462 & 0.4146 & 0.5640 & 0.4920 & 0.5192 & \underline{\ {0.6550}} & \textbf{\textbf{0.6595}} \\
 $\heartsuit$ MUSIQ~\cite{musiq} \upcolor{$\uparrow$} & 74.97 & 74.62 & 72.61 & 59.22 & 61.92 & 75.38 & 70.87 & 71.46 & \textbf{\textbf{76.39}} & \underline{\ {76.04}} \\


 \midrule
  \rowcolor{gray!20}\multicolumn{11}{c}{\ {\textit{CelebRef-HQ-Test}}}\\
\midrule
$\spadesuit$ IDS~\cite{deng2019arcface} \upcolor{$\uparrow$} & 0.3631 & 0.3756 & 0.3422 & 0.3168 & 0.3238 & 0.3372 & 0.3510 & \underline{0.3836} & 0.3540 & \textbf{0.3981} \\
$\spadesuit$ FID~\cite{heusel2017gans} \upcolor{$\downarrow$} & 24.61 & 23.16 & \textbf{21.74} & 47.18 & 31.91 & 24.05 & 31.07 & 22.49 & 24.14 & \underline{\ {21.84}} \\
$\spadesuit$ LPIPS~\cite{lpips} \upcolor{$\downarrow$} & 0.3109 & 0.3124 & 0.3081  & 0.3737 & 0.3464 & \underline{\ {0.2872}} & 0.3615 & 0.3184 & 0.3266 & \textbf{\textbf{0.2827}} \\
 $\heartsuit$ HyperIQA~\cite{hyperiqa}  \upcolor{$\uparrow$} & 0.7305 & \underline{\ {0.7435}}  & 0.7080 & 0.6371 & 0.6012 & 0.7317 & 0.6861 & 0.6800 & 0.7239 & \textbf{\textbf{0.7987}} \\
 $\heartsuit$ TopIQ~\cite{topiq} \upcolor{$\uparrow$} & 0.7906 & \underline{\ {0.8161}}  & 0.7740 & 0.6018 & 0.5934 & 0.7783 & 0.7162 & 0.7075 & 0.8029 & \textbf{\textbf{0.8858}} \\
 $\heartsuit$ MANIQA~\cite{maniqa} \upcolor{$\uparrow$} & 0.5618 & 0.5619 & 0.5201& 0.4456 & 0.4066 & 0.5591 & 0.4962 & 0.5009 & \underline{\ {0.6035}}  & \textbf{\textbf{0.6582}} \\
 $\heartsuit$ MUSIQ~\cite{musiq} \upcolor{$\uparrow$} & 74.71 & \underline{\ {76.10}} & 72.31 & 58.77 & 61.39 & 75.05 & 71.46 & 70.57 & 74.84 & \textbf{\textbf{76.40}} \\
 
 \midrule
  \rowcolor{gray!20}\multicolumn{11}{c}{\ {\textit{LFW-Test}}}\\
\midrule
 $\heartsuit$ HyperIQA~\cite{hyperiqa}  \upcolor{$\uparrow$} & \underline{\ {0.7155}} & 0.6792 & 0.6855 & 0.6371 & 0.5802 & 0.6992 & 0.6489 & 0.5143 & 0.6810 & \textbf{\textbf{0.7816}} \\
 $\heartsuit$ TopIQ~\cite{topiq} \upcolor{$\uparrow$} & \underline{\ {0.6637}} & 0.5962 & 0.5966 & 0.6018 & 0.4511 & 0.6040& 0.5272 & 0.4809 & 0.6297 & \textbf{\textbf{0.7641}} \\
 $\heartsuit$ MANIQA~\cite{maniqa} \upcolor{$\uparrow$} & 0.5539 & 0.4950 & 0.5212 & 0.4456 & 0.3938 & 0.5225 & 0.4884 & 0.4932 & \underline{\ {0.5561}} & \textbf{\textbf{0.6469}} \\
 $\heartsuit$ MUSIQ~\cite{musiq} \upcolor{$\uparrow$} & 75.18 & 74.71 & 73.80 & 66.74 & 61.39 & 75.26& 73.41 & 68.29 & \underline{\ {75.80}} & \textbf{\textbf{76.78}} \\
\midrule
\rowcolor{gray!20}\multicolumn{11}{c}{\ {\textit{Wider-Test}}}\\
\midrule
$\heartsuit$ HyperIQA~\cite{hyperiqa}  \upcolor{$\uparrow$} & \underline{0.7331} & 0.6871 & 0.6854 & 0.6058 & 0.6252 & 0.6995 & 0.6829 & 0.6297 & 0.6413 & \textbf{0.7432} \\
$\heartsuit$ TopIQ~\cite{topiq} \upcolor{$\uparrow$}        & \underline{0.7957} & 0.7444 & 0.7317 & 0.5605 & 0.6407 & 0.7476 & 0.7229 & 0.6347 & 0.6983 & \textbf{0.8276} \\
$\heartsuit$ MANIQA~\cite{maniqa} \upcolor{$\uparrow$}      & \underline{0.5552} & 0.4959 & 0.5044 & 0.4073 & 0.4327 & 0.5205 & 0.4991 & 0.4448 & 0.5132 & \textbf{0.5790} \\
$\heartsuit$ MUSIQ~\cite{musiq} \upcolor{$\uparrow$}        & \underline{74.83} & 73.41  & 71.42 & 53.44 & 65.02  & 74.15  & 72.60  & 65.58  & 68.88  & \textbf{75.50} \\
\midrule

\rowcolor{gray!20}\multicolumn{11}{c}{\ {\textit{WebPhoto-Test}}}\\
\midrule
$\heartsuit$ HyperIQA~\cite{hyperiqa}  \upcolor{$\uparrow$} & \underline{0.7403} & 0.7003 & 0.6913 & 0.6223 & 0.6322 & 0.6923 & 0.6949 & 0.6105 & 0.5933 & \textbf{0.7603} \\
$\heartsuit$ TopIQ~\cite{topiq} \upcolor{$\uparrow$}        & \underline{0.7623} & 0.7243 & 0.6992 & 0.5561 & 0.5925 & 0.7029 & 0.6893 & 0.5928 & 0.6116 & \textbf{0.8061} \\
$\heartsuit$ MANIQA~\cite{maniqa} \upcolor{$\uparrow$}      & \underline{0.5366} & 0.5034 & 0.4909 & 0.4150 & 0.4222 & 0.4935 & 0.5008 & 0.4120 & 0.4396 & \textbf{0.5763} \\
$\heartsuit$ MUSIQ~\cite{musiq} \upcolor{$\uparrow$}        & \underline{74.72} & 74.01  & 70.91 & 58.56 & 65.37  & 72.70  & 73.20  & 63.95  & 63.32  & \textbf{74.98} \\
\bottomrule
 
\end{tabular}
}
\label{Tab:restore_q}
\vspace{-10pt}
}
\end{table*}

We compare A$^\text{2}$BFR with representative face restoration methods, including GAN- and Transformer-based approaches (GFPGAN~\cite{wang2021gfpgan}, CodeFormer~\cite{zhou2022towards}, VQFR~\cite{gu2022vqfr}, DAEFR~\cite{tsai2024dual}), and diffusion-based models (DifFace~\cite{yue2022difface}, DR2~\cite{wang2023dr2}, StableSR~\cite{wang2024exploiting}, VSPBFR~\cite{LU2025111312}). We use the official face-specific variant of StableSR and adopt DiT4SR~\cite{duan2025dit4sr} as a canonical DiT-based baseline.
All reference-based metrics are computed between the restored image and the source ground-truth $I^\text{src}_{\text{GT}}$ to evaluate the fundamental BFR capability of each method. Traditional restoration models do not support prompts as inputs and thus cannot be evaluated on attribute alignment as SC and AA. Consequently, we compare them only in terms of restoration quality and fidelity. To ensure fairness, all prompt-conditioned methods (StableSR~\cite{wang2024exploiting}, DiT4SR~\cite{duan2025dit4sr}, and ours) are evaluated using the template prompt ``A photo of a human face''.

\noindent \textbf{Quantitative Comparisons.} As shown in Table~\ref{Tab:restore_q}, A$^\text{2}$BFR achieves the lowest LPIPS and consistently ranks first or second across all no-reference IQA metrics on AttrFace-90K-Test, CelebRef-HQ-Test,  LFW-Test, Wider-Test, and WebPhoto-Test. 
On AttrFace-90K-Test, our method achieves an LPIPS of 0.2679, yielding a 8.7\% relative reduction compared to the second-best result, and reaches the highest HyperIQA and TopIQ scores.
Similar trends are observed on CelebRef-HQ-Test and real-world datasets, confirming the robustness and perceptual fidelity of our approach across different data domains.

\begin{figure*}[tbp]
    \centering
    \includegraphics[width=\textwidth]{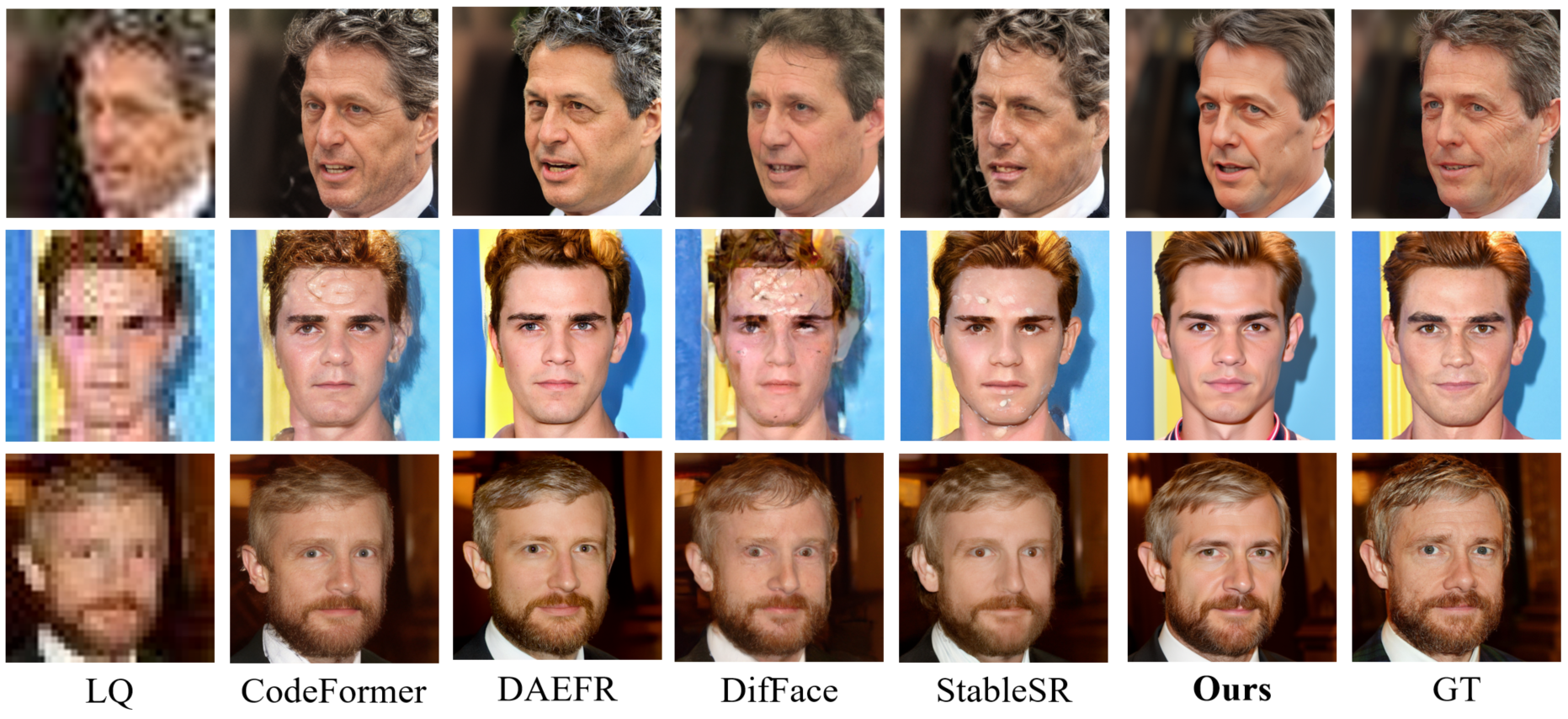}
   \vspace{-1.5em}
    \caption{Qualitative comparison of restoration quality on \textit{CelebRef-HQ-Test}. A$^\text{2}$BFR restores sharper textures and more natural facial details while preserving better fidelity, 
surpassing diffusion- and diffusion-based baselines.} 
    \label{fig:qual_compare}
   \vspace{-0.5em}
\end{figure*}

\noindent \textbf{Qualitative Comparisons.} Fig.~\ref{fig:qual_compare} further illustrates qualitative comparisons on CelebRef-HQ-Test.
While GAN- and diffusion-based methods tend to produce over-smoothed textures or inconsistent local details, A$^\text{2}$BFR restores fine-grained features such as hair and skin details, while preserving fidelity with the input.
Compared with DiT4SR, our reconstructions exhibit clearer facial contours, consistent shading, and more natural local structures around the eyes and mouth.
These results collectively demonstrate that A$^\text{2}$BFR achieves a superior balance between perceptual fidelity and semantic consistency, benefiting from its attribute-aware supervision and semantic dual-training mechanisms.

\subsection{Comparison with Restore-then-edit Methods}

We evaluate attribute-controllable restoration by comparing our single-stage A$^\text{2}$BFR with two-stage restore-then-edit pipelines that pair restoration backbones with SOTA text-driven editors (Step1X-Edit~\cite{liu2025step1x}, Flux.1-Kontext~\cite{labs2025flux}, and Qwen-Image-Edit~\cite{wu2025qwenimagetechnicalreport}).
We include StableSR~\cite{wang2024exploiting} and DiT4SR~\cite{duan2025dit4sr} as restoration baselines because they natively accept prompts, whereas the other methods in Table~\ref{Tab:restore_q} are prompt-agnostic.
For fairness, all compared methods are conditioned on the same target prompt $T^{\mathrm{tar}}$, which specifies the desired attribute change.

\noindent \textbf{Quantitative Comparisons.} Table~\ref{Tab:two-stage} shows that our single-stage A$^\text{2}$BFR achieves state-of-the-art performance in restoration quality, fidelity, and attribute alignment. Single-stage prompt-guided baselines such as StableSR~\cite{wang2024exploiting} exhibit limited controllability, whereas A$^\text{2}$BFR improves AA by +52.58\%. In contrast, two-stage restore--then--edit pipelines suffer substantial fidelity degradation. Compared with the strongest two-stage combination (StableSR~\cite{wang2024exploiting} + Flux.1-Kontext~\cite{labs2025flux}), A$^\text{2}$BFR increases AA by 0.014 while reducing LPIPS by 0.1863, demonstrating better controllability without sacrificing fidelity. A$^\text{2}$BFR also attains the best perceptual quality, achieving the highest HyperIQA score of 0.7908.

\begin{table}[tbp]
  \caption{Quantitative results of attribute alignment comparison on \textit{AttrFace-90K-Test}. \textbf{Best} and \underline{second best} performance are highlighted.}
  \vspace{-1em}
  \centering 
  \resizebox{\columnwidth}{!}{
    \begin{tabular}[t]{l c c c c c c c c c c}
\toprule  
& \multicolumn{2}{c}{\ {Attribute}} & \multicolumn{4}{c}{\ {Fidelity}} & \multicolumn{3}{c}{\ {Quality}} & Human\\
\cmidrule(lr){2-3} \cmidrule(lr){4-7} \cmidrule(lr){8-10}
Methods/Metrics 
& SC \upcolor{$\uparrow$} & AA \upcolor{$\uparrow$}
& LPIPS \upcolor{$\downarrow$} & FID \upcolor{$\downarrow$} & IDS \upcolor{$\uparrow$} & CP-IDS \upcolor{$\uparrow$} & HyperIQA \upcolor{$\uparrow$} & MANIQA \upcolor{$\uparrow$} & MUSIQ \upcolor{$\uparrow$} & Pref. (\%) \\
\midrule

StableSR~\cite{wang2024exploiting}    
 & 92.49 & 0.3278 & \underline{0.3330} & \underline{21.64} & 0.2830 &  \underline{0.7619} & 0.6948 & 0.5178 & 71.35 & 2.4\\

\quad+ Step1X-Edit~\cite{liu2025step1x}        
 & 93.73 & 0.7757 & 0.3889 & 23.93 & 0.2052 & 0.5331 & 0.7491 & 0.6030 & 74.71 & 2.4\\

\quad+ Flux.1-Kontext~\cite{labs2025flux}       
 & \textbf{94.74} & \underline{0.8395} & 0.4726 & 36.36 & 0.1698 & 0.3546 & 0.6152 & 0.4648 & 72.81 & 4.3\\

\quad+ Qwen-Image-Edit~\cite{wu2025qwenimagetechnicalreport}    
 & 94.64 & 0.7490 & 0.4573 & 35.16 & 0.1577 & 0.3059 & 0.5718 & 0.4196 & 73.47 & \underline{10.2} \\

\hdashline

DiT4SR~\cite{duan2025dit4sr}    
 & 93.97 & 0.5714 & 0.3649 & 24.69  & \underline{0.2916} & 0.5630 & 0.7595 & \underline{0.6525} & \textbf{76.33} & 7.3\\

\quad+ Step1X-Edit~\cite{liu2025step1x}       
 & 93.58 & 0.7624 &0.5869 & 25.51 & 0.1983 & 0.5331 & 0.7790 & 0.6478 & 75.51 & 4.7\\

\quad+ Flux.1-Kontext~\cite{labs2025flux}       
 & 94.61 & 0.8214 & 0.4785 & 38.22 & 0.1608 & 0.3763 & \underline{0.6362} & 0.5015 & 74.58 & 4.8\\

\quad+ Qwen-Image-Edit~\cite{wu2025qwenimagetechnicalreport}    
 & 94.54 & 0.7507  & 0.4617 & 34.21 & 0.1638 & 0.3685 & 0.6136 & 0.4707 & 75.76 & 8.8 \\

\hdashline

\rowcolor{gray!20}\textbf{$\text{A}^\text{2}$BFR (Ours)} 
 & \underline{94.70} & \textbf{0.8536} & \textbf{0.2863} & \textbf{21.52} & \textbf{0.3119} & \textbf{0.8301} & \textbf{0.7908} & \textbf{0.6628} & \underline{76.22} & \textbf{55.1} \\

\bottomrule
\end{tabular}
}
\label{Tab:two-stage}
\vspace{-10pt}
\end{table}

We further report Cross-Prompt Identity Similarity (CP-IDS), defined as the ArcFace cosine similarity between two restorations $(\hat I^{\mathrm{src}}_{\mathrm{GT}}, \hat I^{\mathrm{tar}}_{\mathrm{GT}})$ produced from the same LQ input under different prompts. CP-IDS quantifies identity consistency across prompt-conditioned outputs. As shown in Table~\ref{Tab:two-stage}, A$^\text{2}$BFR achieves a CP-IDS of 0.8301,  substantially higher than two-stage pipelines, indicating that it maintains identity consistency under prompt-driven semantic variations. 
This capability also facilitates applications beyond restoration, e.g., constructing identity-preserving face-editing datasets with reliable semantic edits.

\noindent \textbf{User Study.} We conduct a user study to compare the overall subjective preference of A$^\text{2}$BFR and the baselines. As shown in Table~\ref{Tab:two-stage}, our method was selected as the "Best" in \textbf{55.1\%} of the trials regarding the overall preference (covering quality, fidelity, and attribute alignment), far exceeding all other competing methods. More details are provided in the \textit{supplementary materials}.

\begin{figure*}[tbp]
    \centering
    \includegraphics[width=0.9\textwidth]{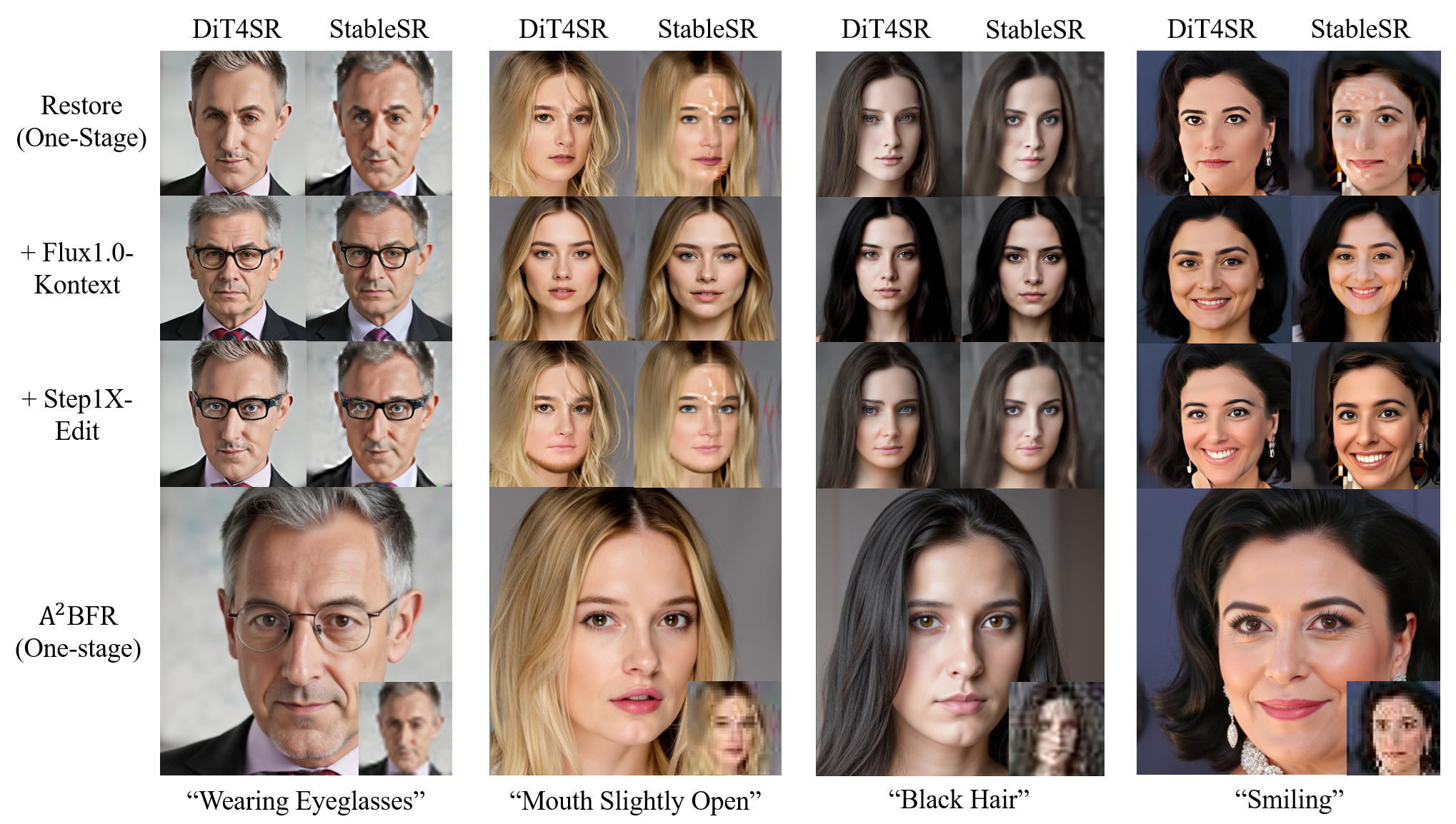}
    \vspace{-2pt}
    \caption{Qualitative comparison of attribute alignment on \textit{AttrFace-90K-Test}. The top row shows one-stage restoration results, while the second and third rows display the results of the two-stage restore-then-edit pipeline. A$^\text{2}$BFR (bottom) achieves better visual quality with accurate attribute control under the provided prompts.}

    \label{fig:attr_compare}
\vspace{-1em}
\end{figure*}


\noindent \textbf{Qualitative Results.} Fig.~\ref{fig:attr_compare} further illustrates that two-stage pipelines often introduce over-edited artifacts or distort facial geometry, whereas A$^\text{2}$BFR preserves fidelity and produces natural, attribute-consistent results. These findings highlight the effectiveness of our AAL and SDT, enabling semantically aligned and perceptually faithful restoration through a single-stage design.

\subsection{Attribute Awareness of $\text{A}^\text{2}$BFR}
\begin{figure*}[tbp]
    \centering
    \includegraphics[width=0.9\textwidth]{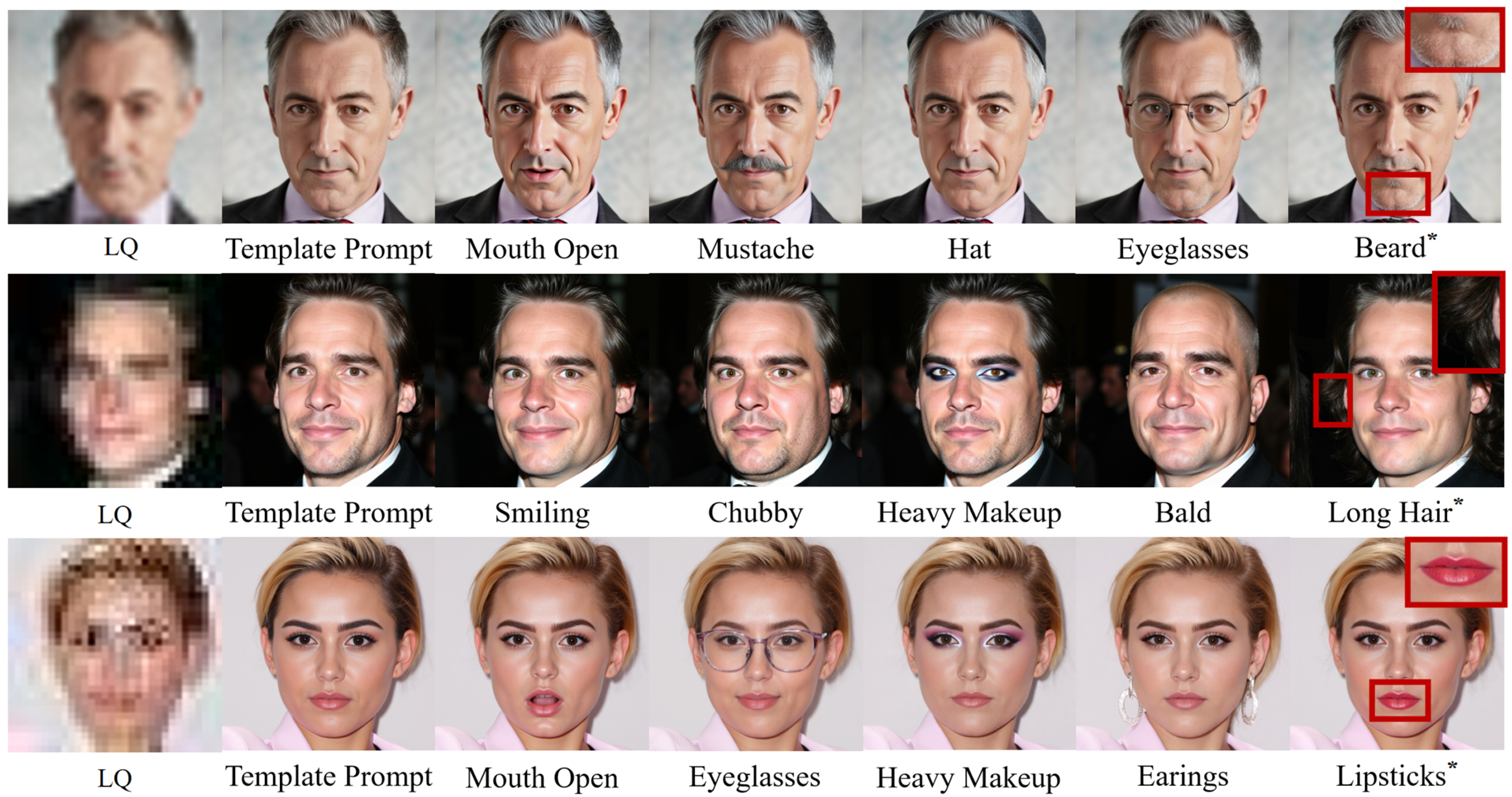}
   \vspace{-5pt}
    \caption{Customizable restoration outputs of A$^\text{2}$BFR. Each column showcases diverse results generated from the same input image by applying different facial attributes, demonstrating the model's ability to perform user-directed BFR. The ``Template Prompt'' presents ``A photo of a human face'', while * denotes out-of-domain attributes.}
    \label{fig:visual2}
\end{figure*}

The inherent attribute awareness of A$^\text{2}$BFR facilitates a paradigm of user-customizable restoration via explicit prompt control. To assess controllability, we perform a qualitative  study: for each fixed LQ face, we apply multiple attribute prompts and visualize the corresponding outputs.
As shown in Fig.~\ref{fig:visual2}, compared with a fixed, attribute-free template prompt, attribute-specific prompts induce clear and consistent changes in the corresponding facial attributes, demonstrating effective prompt control of $\text{A}^\text{2}$BFR.
Overall, these comparisons substantiate that semantic dual-training improves prompt adherence and enables fine-grained attribute editing under the same degraded observation.

A$^\text{2}$BFR exhibits strong generalization beyond the 12 attribute categories defined in AttrFace-90K. As shown in Fig.~\ref{fig:visual2}, it can respond to out-of-domain attribute prompts (e.g., \textit{Beard}, \textit{Lipstick}) and produce the corresponding edits, demonstrating robust cross-attribute generalization. Additional examples are provided in the \textit{supplementary materials}.

\subsection{Ablation and Dataset Analysis}

\noindent \textbf{Effect of AAL and SDT.} Table~\ref{tab:ablation} shows the individual and combined contributions of AAL and SDT with or without $\mathcal{L}_{\text{dual}}$. Incorporating AAL into the baseline substantially improves semantic consistency by 0.21, and attribute accuracy by 0.2136, demonstrating that attribute-guided supervision effectively steers restoration toward the target prompt. Introducing SDT without the dual loss further enhances AA to 0.7757, indicating that exposing the model to semantically paired samples strengthens attribute-specific discrimination.



\begin{table}[tbp]
\centering

\caption{Ablation study of AAL and SDT on \textit{AttrFace-90K-Test}. \textbf{Bold} indicates the best performance and \underline{underline} indicates the second best.}
\vspace{-1em}
\centering 
\setlength{\tabcolsep}{3.5pt}
\resizebox{\columnwidth}{!}{
\begin{tabular}[t]{l c c c c c c c c c }
\toprule  
\multicolumn{3}{c}{\ {Components}} & \multicolumn{2}{c}{\ {Attribute}} & \multicolumn{2}{c}{\ {Fidelity}} & \multicolumn{3}{c}{\ {Quality}}\\
\cmidrule(lr){1-3} \cmidrule(lr){4-5} \cmidrule(lr){6-7} \cmidrule(lr){8-10}
AAL & SDT \textit{w/o} $\mathcal{L}_{\text{dual}}$ & SDT \textit{w} $\mathcal{L}_{\text{dual}}$
& SC \upcolor{$\uparrow$} & AA \upcolor{$\uparrow$}
& LPIPS \upcolor{$\downarrow$} & FID \upcolor{$\downarrow$} & HyperIQA \upcolor{$\uparrow$} & MANIQA \upcolor{$\uparrow$} & MUSIQ \upcolor{$\uparrow$}  \\
\midrule
\xmark & \xmark & \xmark & 94.59 & 0.4078 & 0.2948 & 25.21 & 0.7600 & 0.5892 & 74.43 \\

\cmark & \xmark & \xmark & \textbf{94.80} & 0.6214 & 0.2954 & \underline{22.17} & 0.7561 & 0.6439 & 75.64 \\

\xmark & \cmark & \xmark & 94.63               & 0.6417                & 0.2889                      & 23.76 & 0.7829 & \underline{0.6572} & 75.91 \\

\cmark & \cmark & \xmark & 94.77  & \underline{0.7757}  & \underline{0.2926} & 22.84 & \underline{0.7842} & 0.6505 & \underline{75.94}\\

\rowcolor{gray!20} \cmark & \xmark & \cmark & \underline{94.70}  & \textbf{0.8536} & \textbf{0.2863} & \textbf{21.52}  & \textbf{0.7908} & \textbf{0.6628} & \textbf{76.22} \\
\bottomrule
\end{tabular}
}
\label{tab:ablation}
\vspace{-1 em}
\end{table}

\begin{figure*}[t]
    \centering
    \includegraphics[width=0.9\textwidth]{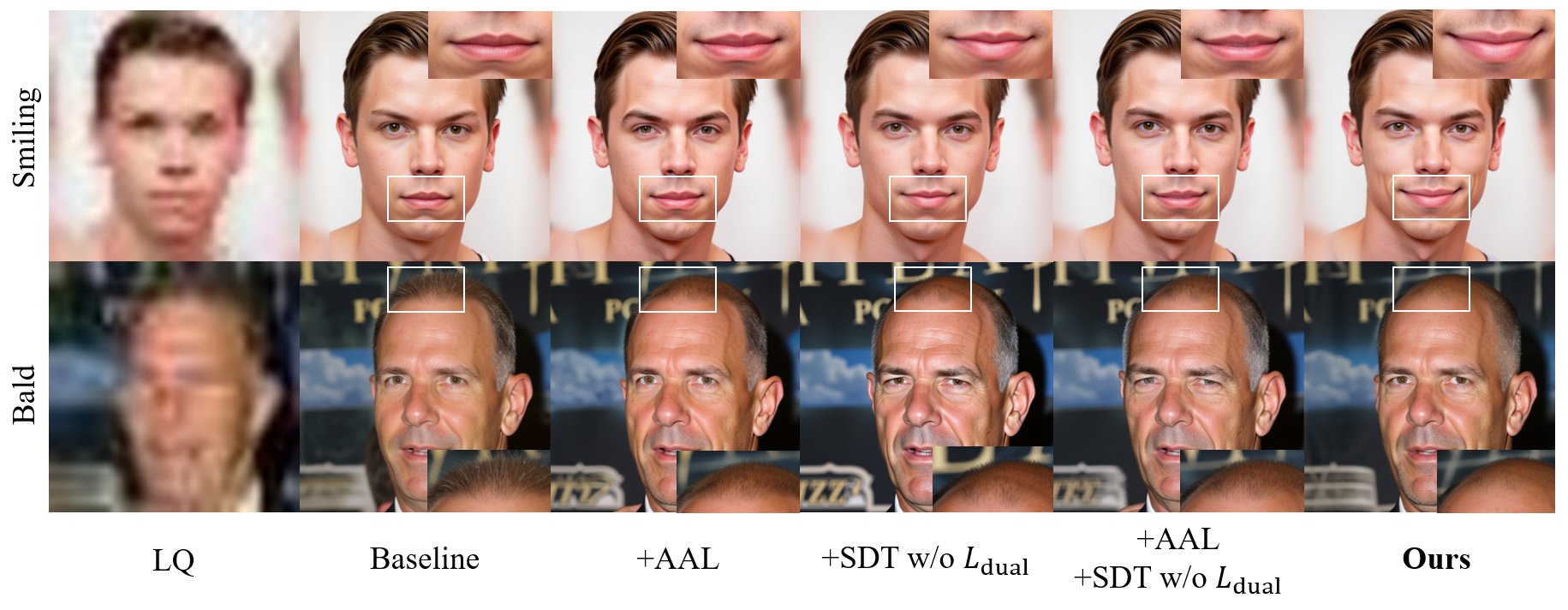}
  \vspace{-0.4em}
    \caption{Qualitative results of ablation study. It can be observed that the responsiveness to attribute texts is progressively enhanced as more components are activated.}
    \label{fig:ablation}
\end{figure*}

Enabling full SDT with $\mathcal{L}_{\text{dual}}$ yields the best result: AA increases to 0.8536, while FID and LPIPS decrease to 21.52 and 0.2863, indicating improved attribute alignment and fidelity. HyperIQA and MANIQA also improve, suggesting that dual-branch semantic supervision complements pixel-level restoration. Qualitative results in Fig.~\ref{fig:ablation} further show that target semantics become more clearly manifested as AAL and SDT are introduced. Overall, AAL injects attribute priors for semantic alignment, SDT promotes prompt-dependent discrimination while maintaining fidelity, and their combination delivers the most controllable and high-quality attribute-guided restoration.

\noindent \textbf{Effect of AttrFace-90K.}
To assess the contribution of the proposed AttrFace-90K dataset, we fine-tune DiT4SR on both FFHQ and AttrFace-90K. As reported in Table~\ref{Tab:dataset_analyse}, fine-tuning on AttrFace-90K yields consistent improvements across most metrics: attribute accuracy increases by +0.1578, semantic consistency by +0.13, and LPIPS decreases by -0.0381, indicating that visual fidelity is preserved. These results verify that AttrFace-90K provides richer attribute-discriminative supervision and generalizes effectively beyond its training distribution, enabling stronger controllable face restoration.

\begin{table}[tbp]
  \caption{Quantitative comparison of face restoration models fine-tuned on FFHQ and AttrFace-90K datasets.}
  \centering
  \vspace{-0.9em}
  \setlength{\tabcolsep}{3.5pt} 
  \small 
  \resizebox{\columnwidth}{!}{
  \begin{tabular}[t]{l c c c c c c c c}
\toprule  
& \multicolumn{2}{c}{\ {Attribute}} & \multicolumn{2}{c}{\ {Fidelity}} & \multicolumn{3}{c}{\ {Quality}}\\
\cmidrule(lr){2-3} \cmidrule(lr){4-5} \cmidrule(lr){6-8}
Methods/Metrics 
& SC \upcolor{$\uparrow$} & AA \upcolor{$\uparrow$} 
& LPIPS \upcolor{$\downarrow$} & FID \upcolor{$\downarrow$} & HyperIQA \upcolor{$\uparrow$} & MANIQA \upcolor{$\uparrow$} & MUSIQ \upcolor{$\uparrow$}  \\
    \midrule
    DiT4SR† (FFHQ) & 94.23  & 0.4730 & 0.3570 & 25.79 & 0.6826 & 0.5283 & 71.33 \\
    DiT4SR† (AttrFace) & \underline{\ {94.36}} & \underline{\ {0.6308}} & \underline{0.3189} & \textbf{21.30} & \underline{0.7069} & \underline{0.5867} & \underline{71.53}\\
    \rowcolor{gray!20}\textbf{$\text{A}^\text{2}$BFR (Ours)} & \textbf{94.70}  & \textbf{0.8536} & \textbf{0.2863} &\underline{21.52}  & \textbf{0.7908} & \textbf{0.6628} &\textbf{76.22}\\
    \bottomrule
  \end{tabular}
  }
 \vspace{-1em}
  \label{Tab:dataset_analyse}
\end{table}

\section{Conclusions}

We propose A$^\text{2}$BFR, an attribute-aware blind face restoration framework that achieves high-fidelity and semantically aligned restoration. The proposed semantic dual-training explicitly enforces prompt-conditioned diversity by requiring distinct outputs for different textual prompts from the same degraded input, while maximizing their semantic distance through an attribute-guided loss. In addition, we introduce AttrFace-90K, a large-scale dataset containing 90K image pairs and 180K fine-grained captions across 12 attributes for supervision and evaluation. Extensive experiments demonstrate that A$^\text{2}$BFR consistently surpasses state-of-the-art methods in perceptual quality, fidelity, and attribute alignment, establishing a controllable paradigm for blind face restoration.



\bibliographystyle{splncs04}
\bibliography{main}
\end{document}